# Assembling Modular, Hierarchical Cognitive Map Learners with Hyperdimensional Computing

Nathan McDonald
Air Force Research Laboratory, Information Directorate
Rome, NY, USA
Nathan.McDonald.5@us.af.mil

Anthony Dematteo
Mohawk Valley Community College
Utica, NY, USA
adematteo13@student.mvcc.edu

*Abstract*— **Cognitive map learners (CML) are a collection of separate yet collaboratively trained single-layer artificial neural networks (matrices), which navigate an abstract graph by learning internal representations of the node states, edge actions, and edge action availabilities. A consequence of this atypical segregation of information is that the CML performs near-optimal path planning between any two graph node states. However, the CML does not learn when or why to transition from one node to another. This work created CMLs with node states expressed as high dimensional vectors consistent with hyperdimensional computing (HDC), a form of symbolic machine learning (ML). This work evaluated HDC-based CMLs as ML modules, capable of receiving external inputs and computing output responses which are semantically meaningful for other HDC-based modules. Several CMLs were prepared independently then repurposed to solve the Tower of Hanoi puzzle without retraining these CMLs and without explicit reference to their respective graph topologies. This work suggests a template for building levels of biologically plausible cognitive abstraction and orchestration.**

## I. Introduction

While classification remains a popular application of deep neural networks (DNN), there is a significant body of research showing that prediction is a critical component of cognition in biological neural networks [1, 2]. Unlike reinforcement learning which seeks to maximize (minimize) an environmental reward (penalty) for an agent [3], this predictive learning is task-agnostic, minimizing instead the error between the actual observed state and the imagined next state given an agent's current state and choice of action. For example, when a calf is born, it spends its first several hours learning not only the movements available to each leg but how to orchestrate them together to walk. Such activities suggests both a compartmentalization and hierarchy of cognition, e.g. the leg movements for walking (motor control) are independent of the choice of direction (goal selection) [21].

However, the DNNs distribute information among all the weights of the network, necessitating a monolithic, end-to-end training algorithm, thereby precluding reuse of previously trained neural subnets in a modular fashion [4]. Ideally, a machine learning (ML) module would be independently optimized for a particular objective yet share some standardized aspect such that it can be arbitrarily integrated (or even be interchangeable) with other ML modules to solve more complex ML problems [4]. By analogy, digital logic has seven fundamental logic gates, each performing a unique function, which can be assembled to solve problems larger than a single Boolean operation, e.g. digital adder, because they share a common information representation: binary.

Predictive learning can be abstracted to learning the topology of a graph, where each node represents a state of being and each edge an action permissible while in that state, e.g. the Tower of Hanoi (ToH) puzzle (Fig. 1a-b). A cognitive map learner (CML) is a collection of single-layer neural networks (matrices) that are separate but collaboratively trained to generate internal representations of the 1) node states, 2) edge actions, and 3) edge action availabilities [5]. Remarkably, though the CML is never explicitly trained for path planning, it can afterwards iteratively compute a near optimal path (fewest edges) between any initial and target node state [5] (Fig. 2a-c).

The CML does not learn when or why to transition from one node state to another; rather, an external source must specify the target node state to start the CML computation. Compartmentalization of information in a CML, by design, permits post-training "brain surgery" to extract the node state representations. This work presents hyperdimensional computing (HDC), or Vector Symbolic Architectures (VSA), [6, 7], as the mathematics for integrating and orchestrating multiple CMLs together as a finite state machine. Instead of learning synaptic weights values, HDC encodes learning by manipulating the similarity among a set of high-dimensional vectors [8, 9, 10]. Being an algebra, such learning may be explicitly expressed in equations that can be edited and reverse engineered, affording both human interpretation and intervention [11]. By working with HDC-compliant hypervectors, information learned by any CML potentially becomes a semantically meaningful input or output to any other HDC-based ML agent, whether an explicit HDC equation or another CML. This work implemented several assemblies and hierarchies of CMLs to simulate movement of rings among pegs to collaboratively solve the ToH puzzle.

The contributions of this work are as follows:

- Specified methods to generate CML node states consistent with HDC hypervector design rules for subsequent symbolic reasoning.

- Designed approaches to uniquely task three CMLs with the same input hypervector.
- Constructed framework of modular, hierarchical CMLs, independently trained and repurposed to solve the Tower of Hanoi puzzle without retraining the CMLs or requiring knowledge of their unique graph topologies.

Section II details the training and operation of a CML and introduces HDC algebra. Section III describes the methods for arbitrarily assembling CMLs. Section IV records the results of this integration, followed by discussion and future applications of this research in Section V.

Matrices are denoted by capital letters and vectors by lower case letters. Importantly, lower case letter vectors come from matrices of the same uppercase letter, e.g. $s_i$ denotes the $i^{th}$ row/column vector of matrix $S$. Key symbols are consolidated and defined in Table I. (Section III.)

## II. BACKGROUND

### A. Cognative Map Learner (CML)

The implementation of CMLs here follows the original paper [5]. Given a graph of $n$ nodes and $e$ edges, a CML with vector lengths $d$ learns three fundamental things: 1) node state representations $S \in \mathbb{R}^{(d,n)}$, 2) edge action representations $A \in \mathbb{R}^{(d,e)}$, and 3) the availabilities $G \in \mathbb{R}^{(e,n)}$ of edge actions from each node state. $S$ and $A$ are both initialized as random Gaussian distributions, $\mu = 0$, with $\sigma = 0.1$ and $\sigma = 1$, respectively. Since the graph topology is known, the gating matrix $G$ is created explicitly as a collection of sparse vectors, one per graph node, with non-zero entries corresponding to edge actions available from each node and zeros everywhere else.

During training, at time $t$, the index of the observed graph node, $o_t$, is translated into the CML's internal node state representation,

$$s_t = S\, o_t, \qquad (1)$$

where $O$ is the identity matrix $I \in \mathbb{R}^{(n,n)}$. The predicted next node state $\hat{s}_{t+1}$ (from the next predicted observable graph node $\hat{o}_{t+1}$) is calculated as the literal sum of the current state $s_t$ and the chosen edge action,

$$\hat{s}_{t+1} = s_t + A\, c_t, \qquad (2)$$

where $C$ is the identity matrix $I \in \mathbb{R}^{(e,e)}$. One training epoch spans all $e$ actions. $S$ and $A$ are updated according to the delta learning rule [12]; whereby, the weights update is calculated as the difference between the actual and the predicted values multiplied by the transpose of the effecting input,

$$\Delta A(t) = \alpha\, (s_{t+1} - \hat{s}_{t+1}) c_t^{\mathsf{T}}, \qquad (3)$$

$$\Delta S(t) = \alpha\, (\hat{s}_{t+1} - s_{t+1}) o_{t+1}^{\mathsf{T}}, \qquad (4)$$

where $\alpha$ is the learning rate, $\alpha = 0.1$. For simplicity, weight updates are summed and applied at the end of each training epoch. Regularization is then applied to preserve unit length among vectors. $S$ is regularized as

$$S = \frac{S}{\|S\|^2}; \qquad (5)$$

while $A$ is regularized along the $e$ axis (column-wise). Training continues until the matrices converge.

The noteworthy feature of CMLs is their path planning ability despite not having been explicitly trained for the task (only pairwise state transitions are presented) (Fig. 2a). An initial observation $o_t$ is specified, defining the current state $s_t$, (1). Then a target observation $o^*$ is provided, producing a target node state,

$$s^* = S\, o^*. \qquad (6)$$

At each time step, the utility of every edge action with respect to $s^*$ is calculated by multiplying the difference between the target and current state by the pseudo-inverse of edge action matrix $A$ [5] (Fig. 2b),

$$u_t = A^{\dagger}\,(s^* - s_t), \qquad (7)$$

where $\dagger$ denotes the Moore-Penrose pseudo-inverse. The gating vector $g_t$ corresponding to node $s_t$ (Fig. 2c) is multiplied elementwise by the utility vector, ensuring only legal edge actions have a positive utility score. A winner-take-all (WTA) algorithm produces a one-hot vector indicating the predicted most useful edge action (Fig. 2d),

$$c_t = WTA(g_t \odot u_t), \qquad (8)$$

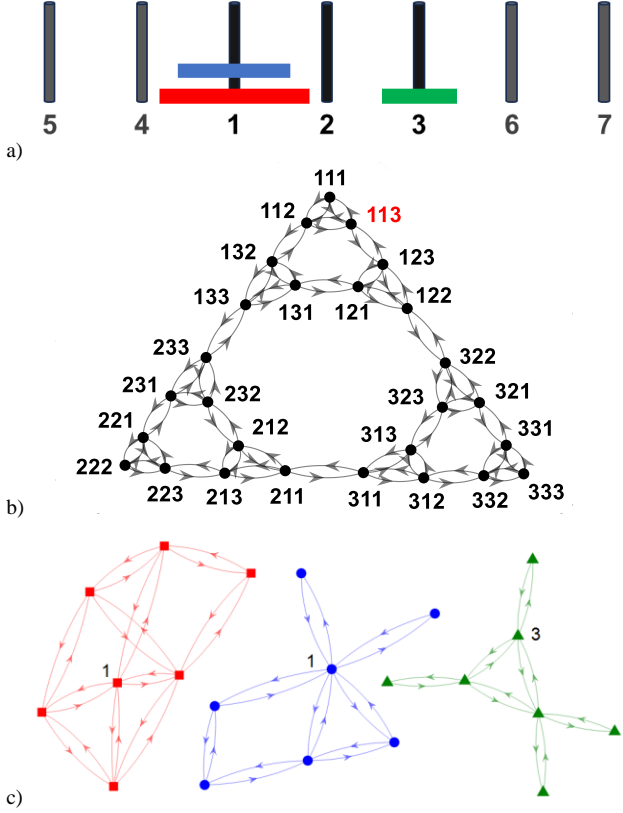

Fig. 1. The Tower of Hanoi (ToH) puzzle game state, $t_{113}$, as described by a) rings on pegs, b) an abstract state graph, and c) a collection of ring states (corresponding to 7 peg positions) from representative graphs for the large (red), medium (green), and small (blue) rings.

where ⊙ denotes elementwise multiplication. The next node state $\hat{s}_{t+1}$ is thus calculated by (2).

Iterating over (7, 8, 2), the CML finds a "reasonable" minimal path between any initial and target state. This solution is competitive with Dijkstra and A* [13], though it lacks the mathematical optimality guarantees [5]. However, the CML provides partial solutions in real-time, rerouting to accommodate for changing target nodes or dropped edges.

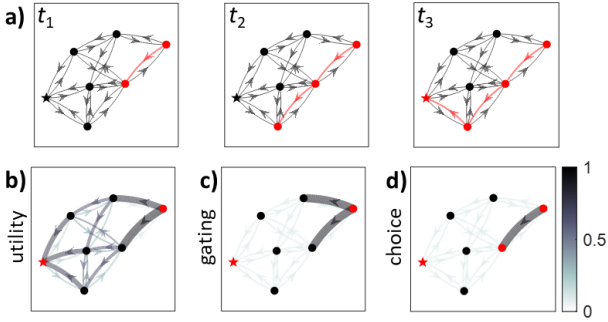

Figure 2. a) CML iteratively path planning (red edges) from its current state (red circle) to the target state (star). Detailing step $t_1$ further, b) the edge action utility values $u$ and c) the gating values $g$ are multiplied together as d) $WTA(g \odot u)$ to determine the choice of new state (red circle). Edge color and width scale with color bar for visual clarity.

### B. Hyperdimensional Computing (HDC)

Instead of artificial neurons and synapses, HDC performs symbolic reasoning with hypervectors, vector of length $d \geq 512$ [6, 8]. As the length of these randomly generated hypervectors increases, the similarity between any two converges to pseudo-orthogonality [6]. Thus, if two symbol hypervectors are not pseudo-orthogonal, then there must be some correlation between them. HDC is an algebra for the creation, manipulation, and measurement of this similarity. These high dimensional relationships are observed for a variety of vector elements, with the community favoring binary {0,1} [8], bipolar {-1,+1} [9], and complex values $e^{j[-\pi,+\pi]}$ [10].

For this work, uniform random bipolar vectors were used according to the Multiplication, Addition, and Permutation (MAP) approach [9]. The following examples assume a dictionary $D$ of known hypervectors, $D = \{w, x, y, z\}$, and a random hypervector $\eta$. Similarity $\delta$ between hypervectors is measured by cosine similarity,

$$\delta(x, y) = \frac{x \cdot y}{\|x\|\|y\|}, \quad (9)$$

where identical vectors have a similarity of 1 and pseudo-orthogonal hypervectors have a similarity close to 0.

The basic operations of HDC are addition, multiplication, permutation, and recovery. Addition and multiplication are elementwise operations, so the dimension of the resultant hypervector remains $d$ regardless of the number of hypervectors added or multiplied together.

Hypervectors may be added together using signed addition, clipping values back to {-1, +1}, e.g.

$$q = sgn([x + y + z]), \quad (10)$$

where

$$sgn(x) := \begin{cases} -1 & \text{if } x < 0 \\ 0 & \text{if } x = 0. \\ 1 & \text{if } x > 0 \end{cases} \quad (11)$$

When adding an even number of hypervectors, a random hypervector $\eta$ is included to break ties. The resulting hypervector is similar to its component hypervectors, $\delta(q, x) \approx \delta(q, y) \approx \delta(q, z) \gg 0$, also called holographic encoding.

Multiplication of hypervectors binds them together, analogous to a key-value pairing. Unlike with addition, the product hypervector is not similar to either of its component hypervectors. Here, the binding operator is elementwise multiplication, which is a self-reversible operation.

Let $q = [w \odot x + y \odot z]$. To approximate the hypervector bound with $w$,

$$\begin{aligned} w \odot q &= w \odot [w \odot x + y \odot z] \\ &= \cancel{w \odot w} \odot x + w \odot y \odot z \\ &= x + \eta \\ &\sim x, \end{aligned} \quad (12)$$

where extraneous terms, being pseudo-orthogonal, are consolidated as a random hypervector $\eta$.

The recovery, or cleanup, operation compares a query hypervector to all other known hypervectors in a dictionary and returns the most similar hypervector above a threshold $\theta$,

$$rec(\sim x, D, \theta) = x. \quad (13)$$

Permutation, denoted $\rho$, is simply a circular shift of the hypervector. This operator is reversible, so it is often used to encode sequences. Let $q = [x + \rho(y) + \rho_2(z)]$,

$$\rho_{-2}(q) \sim z. \quad (14)$$

### III. METHODS

#### A. Accomodating CMLs to HDC

For this work, hypervectors were of length $d = 1e3$. Random hypervectors had maximum similarities $|\delta_{max}(x, y)| < 0.1$, establishing the experimental threshold for pseudo-orthogonality (noise floor) $\theta$. CMLs natively learned pseudo-orthogonal node states $S$, with vector element values normally distributed over the interval [-0.1, +0.1]. Applying the sign operator to a node state, $sgn(s)$, during HDC computation satisfied the bipolar hypervector requirement. Crucially, there remained a high similarity between the actual node states and their respective bipolar versions, $\delta(s_i, sgn(s_i)) = 0.860 \pm 0.030$. CMLs therefore still operates correctly when given a target state $s^* = sgn(s_i)$.

For the reverse case, where one started with a dictionary of bipolar hypervectors for use as $S$, the matrix was simply normalized, (5). Edge actions $A$ were calculated to satisfy (2),

$$a_{ij} = s_i - s_j. \quad (15)$$

For comparison, calculating $A$ produced identical results to training $A$ for the same fixed $S$, (3). As will be shown, CMLs can thus accommodate situations with both arbitrary (learned) and prescribed (calculated) node state hypervectors.

TABLE I. KEY SYMBOLS AND DEFINITIONS

| Category | Symbol | Definition |
|---|---|---|
| CML | $s \in S$ | node state |
| | $a \in A$ | edge action |
| | $g \in G$ | edge action availability |
| | $o$ | observation |
| | $c$ | choice |
| | $s_t$ | current node state |
| | $\hat{s}$ | predicted/prescribed node state |
| | $s^*$ | target node state |
| | $u$ | action utility |
| HDC | $d$ | hypervector length |
| | $\delta$ | cosine similarity |
| | $sgn$ | sign operator |
| | $\odot$ | elementwise multiplication |
| | $\eta$ | random hypervector |
| | $rec$ | recovery |
| | $\rho$ | permutation |
| | $\theta$ | similarity threshold/noise floor when $s^* \in S$ |
| | $\phi$ | similarity threshold when $s^* \approx s_t$ |
| ToH | $t \in T$ | ToH node state |
| | $C_T$ | ToH CML |
| | $l \in L$ | large ring node state |
| | $C_L$ | large ring CML |
| | $m \in M$ | medium ring node state |
| | $C_M$ | medium ring CML |
| | $s \in S$ | small ring node state |
| | $C_S$ | small ring CML |
| | $r \in R$ | ring state vector, viz. $l$, $m$, or $s$ |
| | $C_R$ | ring CML, viz. $C_L$, $C_M$, or $C_S$ |
| | $r_i$ | actual target ring state |
| | $r^*$ | approximate target ring state |
| | $t_{ijk} \in T_\Sigma$ | ToH node state as sum of ring node states |
| | $C_\Sigma$ | composite node state ToH CML |

CMLs as ML modules accepted two inputs: the target state $s^*$ and current state $s_t$, and returned one output: the expected next state $\hat{s}_{t+1}$ (Fig. 3),

$$\hat{s}_{t+1} = C(s^*, s_t). \quad (16)$$

This approach bypasses (1) and (6). Instead, the CML sanities the inputs via node state matrix $S$. If the target state $s^*$ was not in $S$, then it would be pseudo-orthogonal to all node states, $\delta(s^*, S) < \theta$, and the CML returned a zeros vector (the MAP identity element under addition); else if $\delta(s^*, S) \geq \theta$, then the raw $s^*$ vector was used. A cleanup operation, (13), was always applied to the current node state to mitigate state drift errors. Not shown is the termination step. When the target and current node state were sufficiently similar, $\delta(s^*, s_t) \geq \phi$, then the CML returned $s_t$. Unless otherwise stated, $\theta = 0.1$ (noise floor) and $\phi = 0.3$ (experimentally determined).

*B. Orchestrating CMLs*

The Tower of Hanoi (ToH) puzzle is comprised of 3 pegs and 3 rings (Fig. 1a), where the goal is to reconstruct the tower (large, medium, and small ring) on the next peg within the constraints that 1) only one ring may be moved at a time and 2) a larger ring cannot be placed atop a smaller ring. The ToH puzzle is commonly encountered in computer science as an example in recursive programming [14], and previously implemented as spiking neural network (SNN) solution [22]. No such recursion was involved in any of the described solutions.

Under the aforesaid ring movement restrictions, the ToH was modelled as a graph with 27 legal configurations of rings on pegs (node states) and 78 legal moves (edge actions) (Fig. 1b). The baseline ToH CML solution, $C_T$, was created with randomly generated node states, $T \in \mathbb{R}^{(1e3, 27)} \in [-1, +1]$, and calculating $A$, (15), and $G$ (Fig. 4.a).

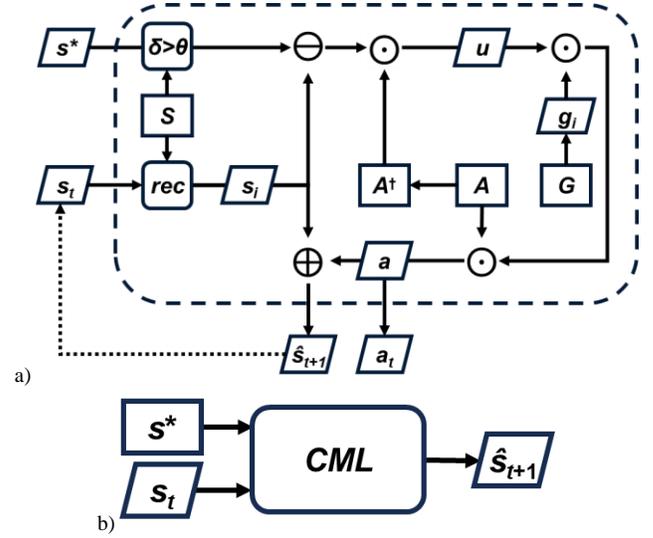

Fig. 3. a) Notional wiring diagram and b) corresponding modular ML symbol for operating (not training) a CML. The CML takes two inputs: target state $s^*$ and current state $s_t$, and returns one output: next state $\hat{s}_{t+1}$. Optional feedback loop shown as dashed line in a). Square boxes indicate static values; parallelagrams indictate changing values.

However, each of the 27 ToH state could be further defined as the collective state of the 3 rings. Indeed, one could consider each ring individually, capable of being on any number of available pegs (graph nodes) (without disambiguating cases of multiple rings on the same peg). Having thus created a CML for each ring without reference to the ToH puzzle itself, how can these ring CMLs be subsequently integrated together to solve the ToH? The four cases described here involve 1) ring CMLs $C_{L,M,S}$ interacting with a global behavioral policy and a global rings state (Fig. 4b), 2) each ring CML interacting with its own local policy and a global rings state (Fig. 4c), 3) the baseline ToH CML $C_T$ dictating the target state which the ring CML implements (Fig. 4d), and 4) constructing a consolidated ToH CML $C_\Sigma$ based on the ring CML states which dictates ring CML target states and accepts the global rings state as a valid current state inputs (Fig. 4e). These myriad approaches illustrate the flexibility afforded by CMLs and HDC for modular ML. In all four cases, there is no retraining of the rings' CMLs or explicit reference to their respective graph topologies.

In the following experiments, each ring was described by a random graph of accessible states ($n = 7$ pegs) (Fig. 1a,c) and governed by a CML trained independently of the other rings. For notational simplicity, an arbitrary ring on peg $i$ is written as node state $r_i$ from matrix $R$. For each ring graph, 3 arbitrary nodes were selected as $r_1$, $r_2$, and $r_3$. While the ToH graph depicted ring state changes as a single edge action, this was not a constraint imposed upon the ring CMLs, such that there could be interstitial node states between $r_1$, $r_2$, and $r_3$. This potential asynchrony may be desirable or detrimental in practice but was ignored here for the sake of illustration.

The first two approaches controlled the ring CMLs by communication with an explicitly computed HDC program,

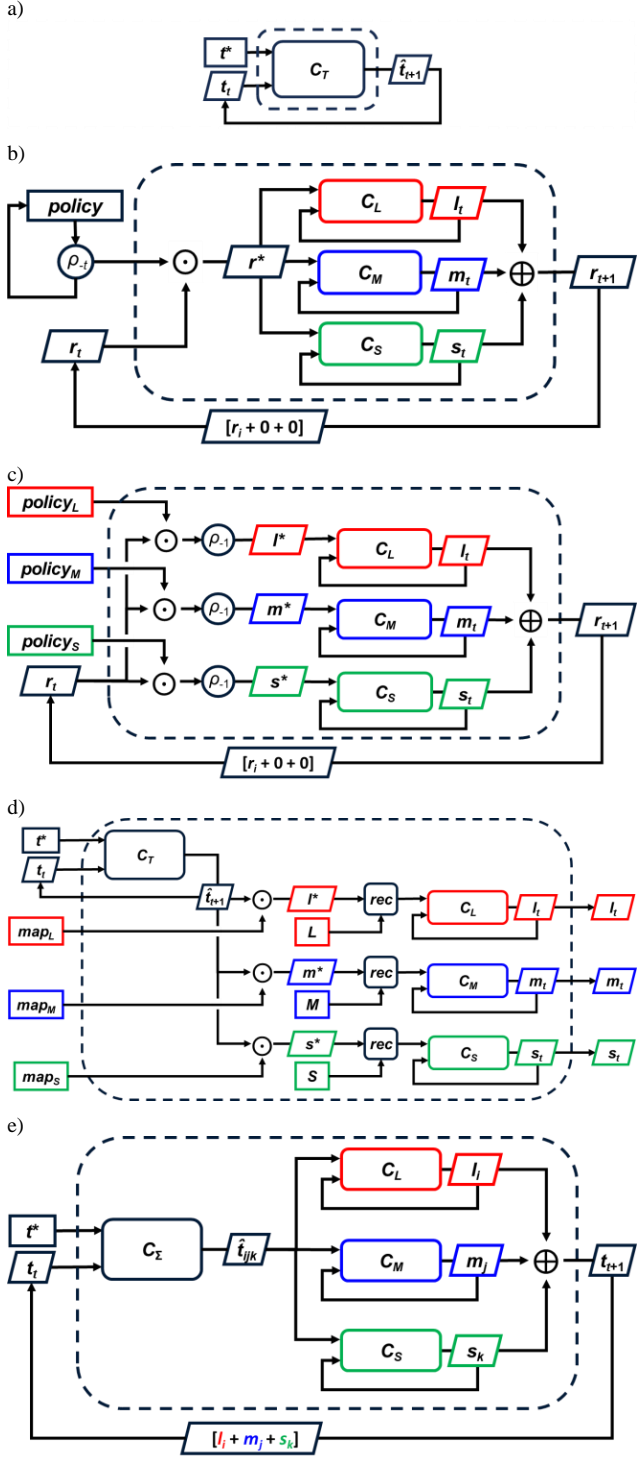

Fig. 4. Wiring diagrams for CML-HDC ToH solutions. Inputs (left) are fed into the ML solution (dashed box) and, except in c), the output (right) becomes the next input. Single ToH solutions are computed via a) an external, monolithic policy and b) a local, partial policies. Arbitrary ToH solutions computed via c) node state mapping and d) using composite node states. Square boxes indicate static values; parallelagrams indictate changing values.

either 1) an external, monolithic policy or 2) local, partial policies. Since only one ring changed state per time step, the sample solution from state 111 → 222 (Fig. 1b) was expressible as a series of stimulus-response pairs [15, 21]. For example, the state transition 111 → 112 → 132, read as "When the small ring moved to peg 2, then the medium ring moved to peg 3," was written as some variation of $s_2 \odot m_3$. If a ring CML did not change its node state, the CML returned the zeros vector, **0**, (not shown in Fig. 3a).

*1) External, monolithic policy:* The sequence of state transitions from 111 → 222 were summed into a single HDC policy hypervector,

$$policy = [s_2 + \rho(s_2 \odot m_3) + \rho_2(m_3 \odot s_3) + \rho_3(s_3 \odot l_2) \\ + \rho_4(l_2 \odot s_1) + \rho_5(s_1 \odot m_2) + \rho_6(m_2 \odot s_2)], \quad (17)$$

where the first $s_2$ was the inciting stimulus and permutations $\rho$ encoded the solution sequence. To start the policy, the initial observed ring state $r_t$ was set to **1**, a ones vector, the multiplicative identity element. For each step in the algorithm (Appendix Table III.), the most recently changed ring state $r_t$ was queried of the HDC policy to retrieve the next target ring state $r^*$ (Fig. 4b),

$$r^* = r_t \odot \rho_{-t}(policy). \quad (18)$$

This target state $r^*$ was broadcasted to each ring CML, but since only one CML recognized $r^*$ as a valid target node, $\delta(r^*, R) > \theta$, only one CML actually changed its current state to $r^*$, which became the next observed ring state $r_{t+1}$. Eq. (17, 15) were then iterated to complete the ToH solution. This approach was self-terminating, since $\rho_{-t}(policy)$ ceased to be semantically meaningful after $t \geq 7$, e.g. $\delta(\rho_{-7}(policy), R) < \theta$.

*2) Local, partial policy:* Instead of communicating with a centralized policy, ring CMLs communicated directly with each other (Fig. 4b). Each ring CML included a local policy with only its ring specific portions of (17),

$$policy_L = [s_3 \odot \rho(l_2)], \quad (19)$$
$$policy_M = [s_2 \odot \rho(m_3) + s_1 \odot \rho(m_2)], \quad (20)$$
$$policy_S = [\rho(s_2) + m_3 \odot \rho(s_3) + l_2 \odot \rho(s_1) + m_2 \odot \rho(s_2)], \quad (21)$$

for the large, medium, and small ring, respectively. By distributing the policy, a global permutation clock was not required. Instead, $\rho$ was applied to the right node state to denote it as the response to the left state stimulus.

To start the policy, the initial ring state $r_t$ was set to **1**. For each time step (Appendix Table III.), the current ring state was queried against each ring's policy,

$$r^* = \rho_{-1}(r_t \odot policy_R); \quad (22)$$

and the appropriate ring CML enacted the state change, becoming the new observable $r_{t+1}$. This method was not self-limiting; however, since $\rho(s_2)$ appeared multiple times in $policy_S$, necessitating an external end condition (not shown).

These two approaches implement a single, explicit ToH solution. Because these methods do not generalize, they will not replicate the general path planning of capability of the ToH CML $C_T$ (Fig. 4a). The next two methods used a ToH CML as the general solution policy which prescribed target ring states to ring CMLs, either 3) by mapping the respective ring states to the ToH CML $C_T$ states or 4) by constructing a ToH CML explicitly from the CML ring states themselves, $C_\Sigma$.

*3) Node state mapping:* The third method mapped the 27 random node states of the ToH CML $C_T$ to each of the 3 ring node states, which were then summed to create mapping hypervectors,

$$map_L = [\sum_j^3 \sum_k^3 t_{1jk} \odot l_1 + \sum_j^3 \sum_k^3 t_{2jk} \odot l_2 + \sum_j^3 \sum_k^3 t_{3jk} \odot l_3], \quad (23)$$

$$map_M = [\sum_i^3 \sum_k^3 t_{i1k} \odot m_1 + \sum_i^3 \sum_k^3 t_{i2k} \odot m_2 + \sum_i^3 \sum_k^3 t_{i3k} \odot m_3], \quad (24)$$

$$map_S = [\sum_i^3 \sum_j^3 t_{ij1} \odot s_1 + \sum_i^3 \sum_j^3 t_{ij2} \odot s_2 + \sum_i^3 \sum_j^3 t_{ij3} \odot s_3]. \quad (25)$$

For example, $t_{1jk}$ represented all 9 ToH node states where the large ring was in the first state (on peg 1).

The ToH CML iteratively calculated the state path from its initial state to the target state, e.g. $t_{111} \rightarrow t_{222}$ (Appendix Table IV.). At each time step, the ToH CML specified the next necessary node state $\hat{t}_{t+1}$, which was broadcast to each ring CML (Fig. 4d). Each ring CML applied its own map to determine the prescribed ring node state,

$$r^* = \hat{t}_{t+1} \odot map_R \quad (26)$$

and iterated through its own CML until attaining the target state as necessary. CML communication in this approach was only one-way, from $C_T$ to the ring CMLs.

*4) Composite node states:* to create the modular, hierarchical ToH CML solution $C_\Sigma$, each ToH node state $t$ was constructed as the sum of the ring node states,

$$t_{ijk} = [l_i + m_j + s_k], \quad (27)$$

such that $\delta(t_{ijk}, l_i) \sim \delta(t_{ijk}, m_j) \sim \delta(t_{ijk}, s_k) = 0.503 \pm 0.023$. Note, these 27 summed node states $t$ were no longer pseudo-orthogonal, having a mean similarity $\delta(t_i, t_j)_{i \neq j} = 0.444 \pm 0.126$, raising the noise floor threshold $\theta$ during subsequent CML operation. The node state matrix $T_\Sigma$ was normalized, (5), before calculating $A_\Sigma$, (15).

For each time step (Appendix Table V.), the $C_\Sigma$ observed current state $t_t$ as the sum of the current ring states (27). $C_\Sigma$ prescribed the next state $\hat{t}_{t+1} = t_{ijk}$. However, in this approach, each ring CML operated upon this query directly (Fig. 4e). Each CML returned their current (new) node state, which were summed to set the actual observed state $t_{t+1}$ for $C_\Sigma$. In this way, $C_\Sigma$ calculated both the ToH puzzle solution and orchestrated the ring CMLs to attain the target ToH state.

## IV. RESULTS

By enforcing CML node states consistent with HDC hypervectors, these states became semantically meaningful to any other HDC-based ML computation. Multiple independently constructed CMLs were arranged post hoc to solve the Tower of Hanoi puzzle. Solutions used 1-4 CMLs, each trained once on their unique graphs without explicit reference to the graph topologies of any other CML.

CMLs were validated according to the fraction of reasonable path solutions over 50 trials of random pairs of initial $s_0$ and target $s^*$ nodes. Only CML solutions with path success = 1.0 were deemed acceptable. All experiments were performed for 50 trials, randomly generating all CML node state matrices and every ring's graph unless stated otherwise.

First, the baseline ToH CML $C_T$ solution attained only 0.846 $\pm$ 0.042 path success (Table II.). The path planning failure cases occurred solely between the node pairs at the sub-triangle corners, viz. nodes (133, 233), (122, 322), and (211, 311) (Fig. 1b). Instead of traversing through the edges to the target state, the CML oscillated between one of these node pairs indefinitely; however, not all paths crossing these edges incurred this failure. Testing instead on the bipolar version of the node states, $t^* = sgn(t)$, corrected this error, producing the expected 1.0 path success for every $C_T$. Composite node state CML $C_\Sigma$ solutions also failed, only attaining path success = 0.831 $\pm$ 0.031, due to the oscillation failure case mentioned above (Table II.). Likewise though, using the bipolar version, $t^* = sgn(t_{ijk})$, resulted in perfect performance.

To disambiguate between any effects from the high degree of symmetry in the ToH graph and the non-orthogonality of the $C_\Sigma$ node states randomly generated graphs of $n = 27$ nodes and $e = 78$ edges were also created (Table II.). For random node states on random graphs, CMLs attained a 1.0 path success. Next, the non-orthogonal $C_\Sigma$ node states were randomly assigned across a random graph; that is, there was no correlation between node proximity and node state similarity. Here too, CMLs attained a 1.0 path success (Table II.). This suggests that graph symmetry is a greater limiting factor in CML performance than non-orthogonality of node states.

TABLE II. SINGLE CML PATH SUCCESS

| Model | $t^*$ | $sgn(t^*)$ |
| --- | --- | --- |
| $C_T$ | 0.846 $\pm$ 0.042 | 1.0 |
| $C_\Sigma$ | 0.831 $\pm$ 0.031 | 1.0 |
| Rand | 1.0 | 1.0 |
| Rand, $[l_i + m_j + s_k]$ | 1.0 | 1.0 |

Moving away from the single CML solutions, the next set of results measured how well different ToH solutions, 1) monolithic, 2) partial, 3) mapping, and 4) composite, orchestrated the three ring CMLs from a single target state hypervector. For the 1) monolithic, 2) partial, and 4) composite methods, the similarity between the supplied approximate target node state $r^*$ and the actual target node state $r_i$ was sufficiently high that the CML continued to operate correctly without an explicit recovery step, $\delta(r^*, r_i) \geq \phi = 0.3$, (Table III.). For 3) mapping, the resulting target states were too noisy to task the ring CMLs directly (path success = 0.889 $\pm$ 0.015, $\theta = \phi = 0.07$). However, the approximate target states were still sufficiently above the noise floor that a recovery operation successfully returned the clean node states, $r^* = r_i$, (Fig. 4d), thereby enabling perfect performance (Table III.).

TABLE III. RING CML PATH SUCCESS

| Model | $r^*$ | $\delta(r^*, r_i)$ | Path Success | | |
|---|---|---|---|---|---|
| | | | $r^*$ | $sgn(r^*)$ | $rec(r^*)$ |
| 1) Monolithic | $r_t \odot \rho_{-t}$ (policy) | 0.308 ± 0.014 | 1.0 | 1.0 | 1.0 |
| 2) Partial | $\rho_{-1}(r_t \odot policy_R)$ | 0.617 ± 0.210 | 1.0 | 1.0 | 1.0 |
| 3) Mapping | $t_{t+1} \odot map_R$ | 0.155 ± 0.032 | 0.889 ± 0.015 | 0.694 ± 0.027 | 1.0 |
| 4) Composite | $t_{ijk}$ | 0.503 ± 0.023 | 1.0 | 1.0 | 1.0 |

## V. DISCUSSION

CMLs are an exciting candidate for modular ML, where modules are independently optimized and arbitrarily (interchangeably) arranged. CMLs can be trained or explicitly calculated, providing both ML and engineering approaches, respectively, to path planning or state transition tasks more generally. Knowledge segregation and column-wise information storage in CMLs enabled post-training "brain surgery" of node states. By conforming these states to HDC hypervector standards, these node states can be subsequently computed upon as semantically meaningful inputs to other HDC-based modules. Since HDC is an algebra, the generated equations are more readily human interpretable and editable than the traditionally "black box" inner workings of DNNs. This provides an interesting bio-plausible approach to constructing hierarchies of command resolution. For example, a CML node might represent a single frame in Eadweard Muybridge's photographs of animal locomotion [16], e.g. a fish "swim right" node state decomposes into a collection of CML node states describing side fins, tail fin, and musculature.

In this work, an environment of three rings among seven pegs was created (Fig. 1a). For each ring, a CML was created, describing arbitrary placement rules local to the ring itself (Fig. 1c), but without respect to the placement of other rings or ToH movement rules. At a later time, there is a request to construct an ML agent to solve the ToH puzzle when the three rings are in a legal configuration among pegs 1, 2, and 3 (Fig. 1b). After extracting the nine relevant ring node states, four methods for implementing ToH solutions were demonstrated (Fig. 4b-e), without any modification to the ring CMLs or knowledge of any their underlying graph structures. In each case, a single hypervector per time step was sufficient to prescribe the next target ring state for the system, though multiple ring-specific policies or maps might be necessary to interpret these commands. HDC's similarity metric natively provided the CMLs with a mechanism for ignoring spurious or irrelevant commands (monolithic and partial policies). At the same time, the holographic nature of addition allowed for multiple ring node states to be simultaneously tasked to unique states with the same target hypervector (mapping and composite).

Canonical solutions to the ToH puzzle recursively implement the following algorithm: 1) Find the largest ring that is not in the correct position; 2) if a smaller ring prevents movement of this larger ring, then move the smaller ring to another legal peg; 3) repeat [23]. This algorithm was successfully implemented in a spatial sematic pointers (SSP) framework, another variant of HDC, using 160,000 leaky integrate and fire neurons (LIF) [22]. SSPs semantically represented the rings, pegs, and sequential goals as spike time patterns processed by several discrete neuroscience-inspired modules: 1) The cortex module encoded 19 context-specific algorithm rules, 2) the basal ganglia module chose the best action, and 3) the thalamus module implemented the selected action. Like the monolithic and local policy methods described above, this SSP approach implemented a single solution of the ToH problem and so, by design, is not as general purpose as the CML framework. However, methods for efficiently implementing HDC operations in spiking neurons are a key aspect of future work in this field [24].

Additional work will focus on HDC-encoded sensor input for interacting with CMLs. Instead of monolithically training DNNs to classify via one-hot vectors, methods that focus on learning general purpose feature extractors, in effect, turn DNNs themselves into modular ML components [25]. Functioning as an ML "analog to digital convertor" (A2D), such feature extractors convert raw data to semantically meaningful hypervectors. For example, the Constrained-Few Shot Class Incremental Learning (C-FSCIL) framework used a pre-trained (and frozen) ResNet-12 feature extractor to populate a dictionary of hypervectors [17]. For the Omniglot dataset, this framework sequentially learned 423 novel classes on top of 1200 base classes while losing less than 1.6% accuracy. HDC augments DNNs instead of replacing them. Further, the use of explicit knowledge dictionaries anticipates collaborative learning, where multiple ML agents can learn diverse things, yet because they share similar semantic dictionaries, they may collaboratively build and share knowledge graphs [18].

Lastly, instead of nodes defining states of being, node states might also be HDC equations, e.g. (17), which are themselves executed sequentially. For example, one task in the MiniGrid environment [19] is for an agent in a maze to pick up a box which is behind a locked door. So the agent must 1) find the key, 2) pick up the key, 3) find the door, 4) unlock the door, 5) find the box, and 6) pick up the box. "*Find x*" and "*pick up x*" are recurring programs, and HDC multiplication readily expresses role-filler pair relationships, e.g. $x \odot key$. One CML might plan the sequence of sub-goals; another might serve as the repository of HDC programs as node states, while lower level HDC-based ML agents actually navigate the maze and perform the specified actions [20].

## VI. CONCLUSION

Cognitive map learners (CML) are a collection of separate yet collaboratively trained ANNs, which learn to traverse an abstract graph. This work created CMLs with graph node states expressed as high dimensional vectors, with the mathematical properties required for hyperdimensional computing (HDC). Enforcing hypervector node states enabled CML node states to be used directly for symbolic reasoning, permitting both input and output of semantically meaningful information. Assemblies and hierarchies of 1-4 CMLs were constructed to solve the

Tower of Hanoi (ToH) puzzle at different levels of abstraction without modification of the ring CMLs or explicit reference to their respective graph topologies, providing a template for building modular, hierarchical machine learning.

ACKNOWLEDGMENT

Any opinions, findings and conclusions, or recommendations expressed in this material are those of the authors, and do not necessarily reflect the views of the US Government, the Department of Defense, or the Air Force Research Lab. Approved for Public Release; Distribution Unlimited: AFRL-2024-1795

REFERENCES

[1] A. Alink, C. Schwiedrzik, A. Kohler, W. Singer, and L. Muckli, "Stimulus predictability reduces responses in primary visual cortex," Journal of Neuroscience, vol. 30, no. 8, 2010, pp. 2960-2966

[2] F. De Lange, M. Heilbron, and P. Kok, "How do expectations shape perception?" Trends in cognitive sciences, vol. 22, no. 9, 2018, pp. 764-779

[3] Y. Duan, X. Chen, R. Houthooft, J. Schulman, and P. Abbeel, "Benchmarking deep reinforcement learning for continuous control," International conference on machine learning, PMLR, 2016, pp. 1329-1338

[4] S. Duan, S. Yu, and J. Príncipe, "Modularizing deep learning via pairwise learning with kernels," IEEE Transactions on Neural Networks and Learning Systems, vol. 33, no. 4, 2021, pp. 1441-1451

[5] Stöckl, Christoph, Yukun Yang, and Wolfgang Maass. "Local prediction-learning in high-dimensional spaces enables neural networks to plan." *Nature Communications* 15, no. ,1 2344, (2024)

[6] D. Kleyko, D. Rachkovskij, E. Osipov, A. Rahimi, "A Survey on Hyperdimensional Computing aka Vector Symbolic Architectures, Part I: Models and Data Transformations," ACM Computing Surveys, vol. 55, is. 6, no. 130, 2023, pp 1–40

[7] D. Kleyko, D. Rachkovskij, E. Osipov, A. Rahimi, "A Survey on Hyperdimensional Computing aka Vector Symbolic Architectures, Part II: Applications, Cognitive Models, and Challenges," ACM Computing Surveys, vol. 55, is. 9, no. 175, 2023, pp 1–52

[8] P. Kanerva, "Hyperdimensional computing: An introduction to computing in distributed representation with high-dimensional random vectors," Cognitive computation, vol. 1, 2009, pp.139-159

[9] R. Gayler, "Multiplicative binding, representation operators & analogy (workshop poster)," (1998).

[10] T. Plate, *Holographic Reduced Representation: Distributed representation for cognitive structures*. Vol. 150. Stanford: CSLI Publications, 2003.

[11] P. Neubert, S. Schubert, and P. Protzel, "An introduction to hyperdimensional computing for robotics," KI-Künstliche Intelligenz, vol. 33, 2019, pp. 319-330

[12] J. Hertz, A. Krogh, and R. Palmer, *Introduction to the theory of neural computation*. CRC Press, 2018.

[13] S. Russell and P. Norvig, Artificial Intelligence: A modern approach, Hoboken, NJ: Pearson. 2021

[14] R. Graham, D. Knuth, and O. Patashnik, Concrete Mathematics. Reading, Mass.: Addison-Wesley, 1989

[15] D. Kleyko, E. Osipov, R. Gayler, A. Khan, and A. Dyer, "Imitation of honey bees' concept learning processes using vector symbolic architectures," Biologically Inspired Cognitive Architectures, vol. 14, 2015, pp. 57-72

[16] E. Muybridge, Animals in motion, Courier Corporation, 2012

[17] M. Hersche, G. Karunaratne, G. Cherubini, L. Benini, A. Sebastian, and A. Rahimi, "Constrained few-shot class-incremental learning," IEEE/CVF Conference on Computer Vision and Pattern Recognition, 2022, pp. 9057-9067

[18] P. Poduval, H. Alimohamadi, A. Zakeri, F. Imani, M. Najafi, T. Givargis, and M. Imani, "Graphd: Graph-based hyperdimensional memorization for brain-like cognitive learning," *Frontiers in Neuroscience* 16, 2022, 757125.

[19] M. Chevalier-Boisvert, B. Dai, M. Towers, R. de Lazcano, L. Willems, S. Lahlou, S. Pal, P. Castro, and J. Terry, "Minigrid & Miniworld: Modular & Customizable Reinforcement Learning Environments for Goal-Oriented Tasks," arXiv preprint arXiv:2306.13831, 2023

[20] B. Komer, T. Stewart, A. Voelker, and C. Eliasmith, "A neural representation of continuous space using fractional binding," In *CogSci*, 2019, pp. 2038-2043.

[21] D. Rachkovskij, A., Ernst M. Kussul, and Tatiana N. Baidyk. "Building a world model with structure-sensitive sparse binary distributed representations." *Biologically Inspired Cognitive Architectures* 3, 64-86, 2013

[22] T. Stewart, and C. Eliasmith. "Neural cognitive modelling: A biologically constrained spiking neuron model of the Tower of Hanoi task." In *Proceedings of the Annual Meeting of the Cognitive Science Society*, vol. 33, no. 33. 2011.

[23] H. Simon, "The functional equivalence of problem solving skills," *Cognitive Psychology*, 7(2), 268–288, 1975

[24] J. Orchard and Russell Jarvis. "Hyperdimensional Computing with Spiking-Phasor Neurons." In *Proceedings of the 2023 International Conference on Neuromorphic Systems*, pp. 1-7. 2023.

[25] S. Duan, "Unifying Kernel Methods and Neural Networks and Modularizing Deep Learning," PhD diss., University of Florida, 2020.

APPENDIX

For visual clarity, only the relevant portions of *policy* or *map* hypervectors are shown for each time step; extraneous terms are consolidated as $\eta$. **1** denotes ones vector, **0** denotes zeros vector.

TABLE IV. MONOLITHIC & PARTIAL POLICIES

| Time Step | Board State | ToH State | 1) Monolithic $r^* = r_t \odot \rho_{-t}$ (policy) | 2) Partial $r^* = \rho_{-1}(r_t \odot policy_R)$, | $r_{t+1} = \sum C_R(r^*, r_t)$ |
|---|---|---|---|---|---|
| 0 | 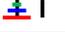 | 111 | $\mathbf{1} \odot \rho_0[s_2 + \eta]$ | $\mathbf{1} \odot policy_S$ $= \mathbf{1} \odot \rho_{-1}[\rho(s_2) + \eta]$ | $[\mathbf{0} + \mathbf{0} + s_2] = s_2$ |
| 1 | 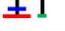 | 112 | $s_2 \odot \rho_{-1}[\rho(s_2 \odot m_3) + \eta]$ | $s_2 \odot policy_M$ $= s_2 \odot \rho_{-1}[s_2 \odot \rho(m_3) + \eta]$ | $[\mathbf{0} + m_3 + \mathbf{0}] = m_3$ |
| 2 | 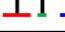 | 132 | $m_3 \odot \rho_{-2}[\rho_2(m_3 \odot s_3) + \eta]$ | $m_3 \odot policy_S$ $= m_3 \odot \rho_{-2}[m_3 \odot \rho_2(s_3) + \eta]$ | $[\mathbf{0} + \mathbf{0} + s_3] = s_3$ |
| 3 | 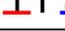 | 133 | $s_3 \odot \rho_{-3}[\rho_3(s_3 \odot l_2) + \eta]$ | $s_3 \odot policy_L$ $= s_3 \odot \rho_{-3}[s_3 \odot \rho_3(l_2) + \eta]$ | $[l_2 + \mathbf{0} + \mathbf{0}] = l_2$ |
| 4 | 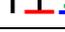 | 233 | $l_2 \odot \rho_{-4}[\rho_4(l_2 \odot s_1) + \eta]$ | $l_2 \odot policy_S$ $= l_2 \odot \rho_{-4}[l_2 \odot \rho_4(s_1) + \eta]$ | $[\mathbf{0} + \mathbf{0} + s_1] = s_1$ |
| 5 | 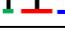 | 231 | $s_1 \odot \rho_{-5}[\rho_5(s_1 \odot m_2) + \eta]$ | $s_1 \odot policy_M$ $= s_1 \odot \rho_{-5}[s_1 \odot \rho_5(m_2) + \eta]$ | $[\mathbf{0} + m_2 + \mathbf{0}] = m_2$ |
| 6 | 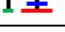 | 221 | $m_2 \odot \rho_{-6}[\rho_6(m_2 \odot s_2) + \eta]$ | $m_2 \odot policy_S$ $= m_2 \odot \rho_{-6}[m_2 \odot \rho_6(s_2) + \eta]$ | $[\mathbf{0} + \mathbf{0} + s_2] = s_2$ |
| 7 | 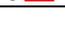 | 222 | $s_2 \odot \rho_{-7}[policy]$ | END | $[\mathbf{0} + \mathbf{0} + \mathbf{0}] = \mathbf{0}$ |

TABLE V. MAPPING & COMPOSITE HIERARCHIES

| Time Step | $t_t$ | $\hat{t}_{t+1}$ | 3) Mapping $r^* = \hat{t}_{t+1} \odot map_R$ | $C_R(r^*, r_t)$ | $\hat{t}_{ijk} = r^*$ | 4) Composite $C_R(r^*, r_t)$ | $t_{t+1} = \sum C_R(r^*, r_t)$ |
|---|---|---|---|---|---|---|---|
| 0 | $t_{111}$ | $t_{112}$ | $s^* = t_{112} \odot [t_{112} \odot s_2 + \eta]$ | $l_1, m_1, \mathbf{s_2}$ | $t_{112} = [l_1 + m_1 + s_2]$ | $C_S(t_{112}, s_1) = s_2$ | $t_{112} = [l_1 + m_1 + \mathbf{s_2}]$ |
| 1 | $t_{112}$ | $t_{132}$ | $m^* = t_{132} \odot [t_{132} \odot m_3 + \eta]$ | $l_1, \mathbf{m_3}, s_2$ | $t_{132} = [l_1 + m_3 + s_2]$ | $C_M(t_{132}, m_1) = m_3$ | $t_{132} = [l_1 + \mathbf{m_3} + s_2]$ |
| 2 | $t_{132}$ | $t_{133}$ | $s^* = t_{133} \odot [t_{133} \odot s_3 + \eta]$ | $l_1, m_3, \mathbf{s_3}$ | $t_{133} = [l_1 + m_3 + s_3]$ | $C_S(t_{133}, s_2) = s_3$ | $t_{133} = [l_1 + m_3 + \mathbf{s_3}]$ |
| 3 | $t_{133}$ | $t_{233}$ | $l^* = t_{233} \odot [t_{233} \odot l_2 + \eta]$ | $\mathbf{l_2}, m_3, s_3$ | $t_{233} = [l_2 + m_3 + s_3]$ | $C_L(t_{233}, l_1) = l_2$ | $t_{233} = [\mathbf{l_2} + m_3 + s_3]$ |
| 4 | $t_{233}$ | $t_{231}$ | $s^* = t_{231} \odot [t_{231} \odot s_1 + \eta]$ | $l_2, m_3, \mathbf{s_1}$ | $t_{231} = [l_2 + m_3 + s_1]$ | $C_S(t_{231}, s_3) = s_1$ | $t_{231} = [l_2 + m_3 + \mathbf{s_1}]$ |
| 5 | $t_{231}$ | $t_{221}$ | $m^* = t_{221} \odot [t_{221} \odot m_2 + \eta]$ | $l_2, \mathbf{m_2}, s_1$ | $t_{221} = [l_2 + m_2 + s_1]$ | $C_M(t_{221}, m_3) = m_2$ | $t_{221} = [l_2 + \mathbf{m_2} + s_1]$ |
| 6 | $t_{221}$ | $t_{222}$ | $s^* = t_{222} \odot [t_{222} \odot s_2 + \eta]$ | $l_2, m_2, \mathbf{s_2}$ | $t_{222} = [l_2 + m_2 + s_2]$ | $C_S(t_{222}, s_1) = s_2$ | $t_{222} = [l_2 + m_2 + \mathbf{s_2}]$ |